\documentclass[11pt,a4paper]{article}
\usepackage[hyperref]{acl2019}
\usepackage{times}
\usepackage{latexsym}
\usepackage{graphicx}
\usepackage{url}
\usepackage{multirow}
\usepackage{todonotes}
\usepackage{algorithm}
\usepackage{algorithmic}
\usepackage{amsmath}
\usepackage{enumerate}
\usepackage{balance}

\usepackage{booktabs}
\usepackage{bm}
\usepackage{amssymb}
\usepackage{amsfonts}
\usepackage{mathtools}
\usepackage{comment}
\usepackage{subfigure}

\usepackage{color}
\newcommand{\red}[1]{\textcolor{red}{#1}}
\newcommand{\blue}[1]{\textcolor{blue}{#1}}

\usepackage{array}
\newcolumntype{L}[1]{>{\raggedright\arraybackslash}p{#1}}
\newcolumntype{R}[1]{>{\raggedleft\arraybackslash}p{#1}}
\newcolumntype{C}[1]{>{\centering\let\newline\\\arraybackslash\hspace{0pt}}m{#1}}

\aclfinalcopy 
\def\aclpaperid{2766} 

\title{Target-Guided Open-Domain Conversation}

\author{Jianheng Tang$^{1,2}$,~~ Tiancheng Zhao$^{2}$,~~ Chenyan Xiong$^{3}$,~~ Xiaodan Liang$^1$\thanks{~~corresponding authors}, \\ 
{\bf Eric P. Xing}$^{2,4}${\bf ,}~~ {\bf Zhiting Hu}$^{2,4*}$ \\
{\small $^1$Sun Yat-sen University, $^2$Carnegie Mellon University, $^3$Microsoft Research AI, $^4$Petuum Inc.} \\
{\small\texttt{\{sqrt3tjh,xdliang328\}@gmail.com, \{tianchez,zhitingh\}@cs.cmu.edu}}\\
{\small\texttt{chenyan.xiong@microsoft.com,  eric.xing@petuum.com}}
}
\date{}

\begin{document}
\maketitle

\begin{abstract}
Many real-world open-domain conversation applications have specific goals to achieve during open-ended chats, such as  recommendation, psychotherapy, education, etc. We study the problem of imposing conversational goals on open-domain chat agents. In particular, we want a conversational system to chat naturally with human and \emph{proactively guide} the conversation to a designated target subject. The problem is challenging as no public data is available for learning such a target-guided strategy. We propose a structured approach that introduces coarse-grained keywords to control the intended content of system responses. We then attain smooth conversation transition through turn-level supervised learning, and drive the conversation towards the target with discourse-level constraints. We further derive a keyword-augmented conversation dataset for the study. Quantitative and human evaluations show our system can produce meaningful and effective conversations, significantly improving over other approaches\footnote{Data and code are publicly available at \url{https://github.com/squareRoot3/Target-Guided-Conversation}}.
\end{abstract}

\section{Introduction}\label{sec:intro}

\begin{figure}[t]
\centering
  \includegraphics[width=0.98\linewidth]{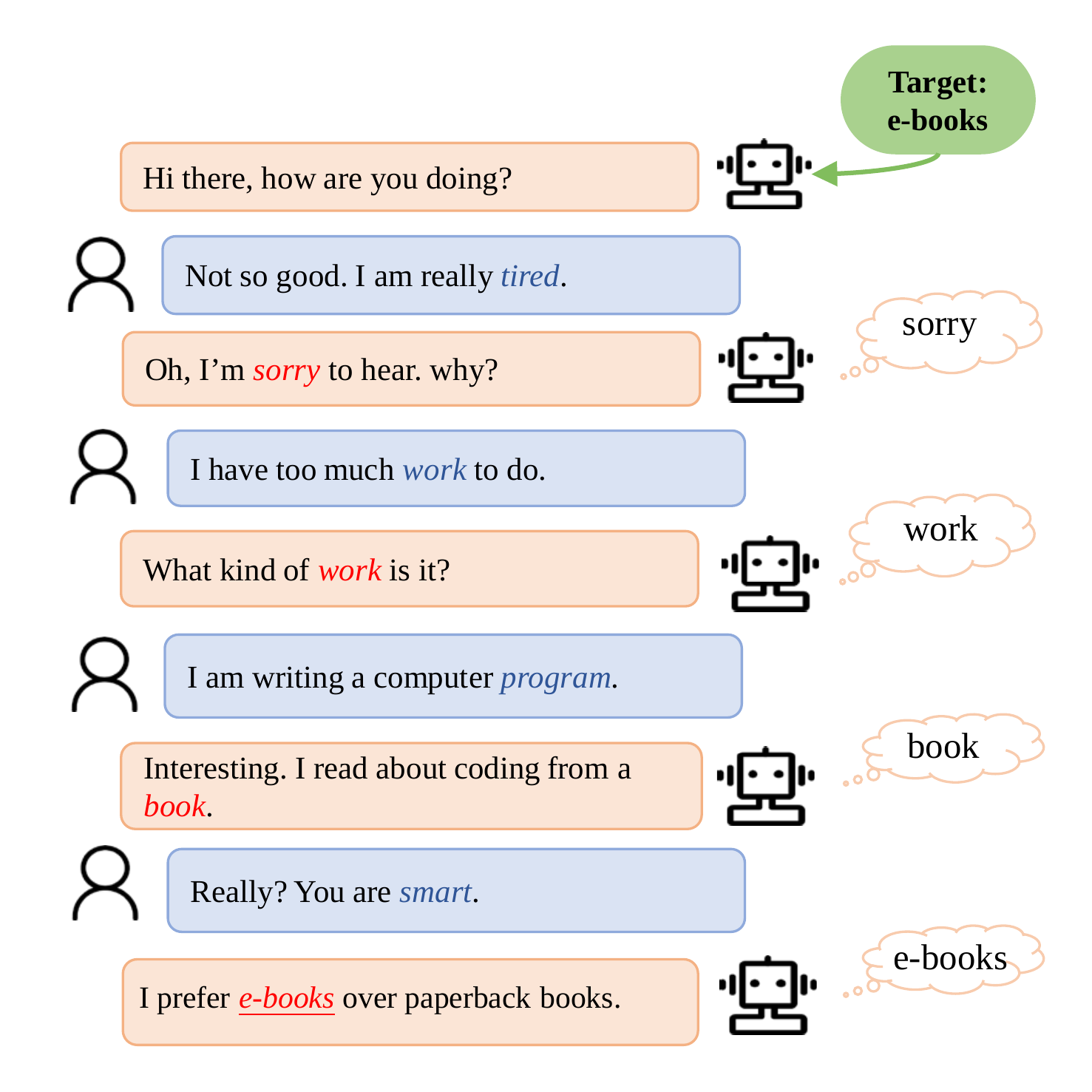}
  \vspace{-10pt}
\caption{
Target-Guided Open-Domain Conversation. The agent is given a target subject \emph{e-books} which is unknown to the human. The goal is to guide the conversation naturally to the target. Utterance keywords are highlighted in {\color{red} red} (agent) and {\color{blue} blue} (human) and in \emph{italic}.
}
\label{fig:intro}
\vspace{-8pt}
\end{figure}

Creating intelligent agent that can carry out open-domain conversation with human is a long-lasting challenge. Impressive progress has been made, advancing from early rule-based systems, e.g., Eliza~\citep{weizenbaum1966eliza}, to recent end-to-end neural conversation models that are trained on massive data~\citep{shang2015neural,li2015diversity} and make use of background knowledge~\citep{fang2018sounding,ghazvininejad2018knowledge,liu2018knowledge}.

However, current open-domain systems still struggle to conduct engaging conversations~\cite{ram2018conversational}, and often generate inconsistent or uncontrolled results. Further, many practical open-domain dialogue applications do have specific goals to achieve even though the conversations are open-ended, e.g., accomplishing nursing goals in therapeutic conversation, inspiring ideas in education, making recommendation and persuasion, and so forth. Thus, there is a strong demand to enable the integration of goals and strategy into open-domain dialogue systems, and it imposes challenges to both: first, how to define the goal for an open-domain chat system; and second, how to encode dialogue strategy into the response production process. 
It is also crucial to attain a general method that is not tailored towards specialized goals that require domain-specific handcrafting and annotations~\citep{yarats2017hierarchical,he2018decoupling,li2018towards}.



This paper makes a step towards open-domain dialogue agents with conversational goals.
In particular, we want the system to chat naturally with humans on open domain topics and \emph{proactively guide} the conversation to a designated target subject. For example, in Figure~\ref{fig:intro}, given a target \emph{e-books} and an arbitrary starting topic such as \emph{tired}, the agent drives the conversation in a natural way following a high-level logical backbone, and effectively reaches the target in the end. 
Such a target-guided conversation setup is general-purpose and can entail a large variety of practical applications as above. The above problem is difficult in that the agent has to balance well between chatting naturally and achieving the target; and moreover, to the best of our knowledge, there is no public dataset available for learning target-guided dialogue.


This paper proposes a solution to the task. We decouple the whole system into separate modules and address the challenges at different granularity. Specifically, we explicitly model and control the intended content of each system response by introducing coarse-grained utterance keywords. We then impose a discourse-level rule that encourages the keywords to approach the end target during the course of the conversation; and we attain smooth conversation transition at each dialogue turn through turn-level supervised learning. To this end, we further derive a keyword-augmented conversation dataset from an existing daily-life chat corpus~\citep{zhang2018personalizing} and use it for learning keyword transitions and utterance production.


We study different keyword transition approaches, including pairwise PMI-based transition, neural-based prediction, and a hybrid kernel-based method. We conduct quantitative and human evaluations to measure the performance of sub-modules and the whole system. Our agent is able to generate meaningful and effective conversations with a decent success rate of reaching the targets, improving over other approaches in different respects. We show target-guided open-domain conversation is a promising and potentially important direction for future research.

\section{Related Work}

The past end-to-end dialogue research can be broadly divided into two categories: task-oriented dialogue systems and chat-oriented (a.k.a open-domain) systems. For task-oriented dialogue systems, the system is designed to accomplish specific goals, e.g., providing bus schedule~\cite{raux2005let,young2007hidden,dhingra2017towards}. Besides information giving, other tasks have been extensively studied, such as negotiations~\cite{devault2015toward,lewis2017deal,he2018decoupling,cao2018emergent}, symmetric collaborations~\cite{he2017learning}, etc. 

On the other hand, chat-oriented dialogue systems have been created to model open-domain conversations without specific goals. Prior work has been focusing on developing novel neural architectures that improve next utterance generation or retrieval task performance by training on large open-domain chit-chat dataset~\cite{sordoni2015neural,serban2016hierarchical,Zhou2016MultiviewRS,Wu2018MultiTurnRS}. However, despite the steady improvement over model architectures, 
the current systems can still suffer from a range of limitations, e.g., dull responses, inconsistent persona~\cite{li2016persona}, etc. 

The commercial chatbot XiaoIce~\cite{zhou2018design} and the first Amazon Alexa challenge winner~\cite{fang2018sounding} have stressed to improve engagement with users. 
Also, to encourage discourse-level strategy, prior work has developed different system action representations that enable the model to reason at the dialogue level. One line of work has utilized latent variable models~\cite{zhao2017learning,yarats2017hierarchical,zhao2019rethinking} to infer a latent representation of system responses, which separates the natural language generation process from decision-making. Another approach has created hybrid systems to incorporate hand-crafted coarse-grained actions~\cite{williams2017hybrid,he2018decoupling} as a part of the neural dialogue systems. These systems have typically focused on specific domains such as price negotiation and movie recommendation.
Building upon the prior work, this paper creates novelty in terms of both defining goals for open-domain chatting and creating system actions representations.
Our structured solution use predicted keywords as a non-parametric representation of the intended content for the next system response.

Due to the lack of full supervision data, the solution proposed in this work divides the task into two competitive sub-objectives, each of which can be conquered with either direct supervision or simple rules. Such a divide-and-conquer approach represents a general means of addressing complex task objectives with no end-to-end supervision available. A similar approach has been adopted in other contexts, such as text style transfer~\citep{hu2017toward,shen2017style,yang2018unsupervised} and content manipulation~\citep{wang2019toward}, where content fidelity as a sub-objective is achieved with simple auto-encoding training, while the competitive nature of multiple sub-objectives jointly drives the models to learn desired behaviors.


\section{Task Definition: Target-guided Open-domain Conversation}\label{sec:task}
We first formally define the task of target-guided open-domain conversation. We also establish the key notations used in the rest of the paper.

Briefly, given a target, we want a chat agent to converse with human starting from an arbitrary initial topic, and lead the conversation to the target in the end. In this paper, we define a target to be a word (e.g., an entity name \emph{McDonald}, or a common noun \emph{book}, etc.) and denote it as $t$. We note that a target can also be formulated in other more complex forms depending on specific applications. The target is only presented to the agent and is unknown to the human. The conversation starts with an initial topic which is usually randomly picked by the human. At each dialogue turn where the agent wants to make a response, it has access to the conversation history consisting of a sequence of utterances by either the human or the agent, $\bm{x}_{1:n}=\{\bm{x}_1, \dots, \bm{x}_n\}$. The agent then produces an utterance $\bm{x}_{n+1}$ as a response, aiming to satisfy (1) {\bf transition smoothness} by making the response natural and appropriate in the current conversation context, and (2) {\bf target achievement} by driving the conversation to reach the designated target. Specifically, we consider a target is achieved when either the human or the agent mentions the target or similar word in an utterance---such a definition is simple and allows easy measurement of the success rate. Again, other more complex and meaningful measures could be considered for specific practical applications.

The above two objectives are complementary and competitive. On one hand, an agent cannot simply bring up the target content regardless of the conversation context. For example, given a target \emph{cat} and conversation history \emph{\{Human: I went to a movie.\}}, a response like \emph{Do you like cat?} is typically not a smooth transition, even though it quickly reaches the target. On the other hand, the agent must avoid being trapped in open-ended chats by producing only smooth yet reactive responses. Instead, it has to proactively lead the conversation to approach the target.

The competitive nature of the two desiderata requires the agent to grasp a conversation strategy that balances well between different factors. 
To the best of our knowledge, there is no public large data that fits the new problem setting and permits end-to-end learning of such a discourse-level strategy in open domain. 
Instead, we usually only have access to those open-ended conversation data where interlocutors conversed freely without a specified end target.

To this end, we propose to break down the problem, leverage partial supervisions and introduce more structures for a solution. In the following, we first present our approach to the task (section~\ref{sec:method}), and then introduce a large open-ended conversation dataset used for building the conversational agent (section~\ref{sec:data}).



\begin{figure*}[t]
\centering
  \includegraphics[width=1\linewidth]{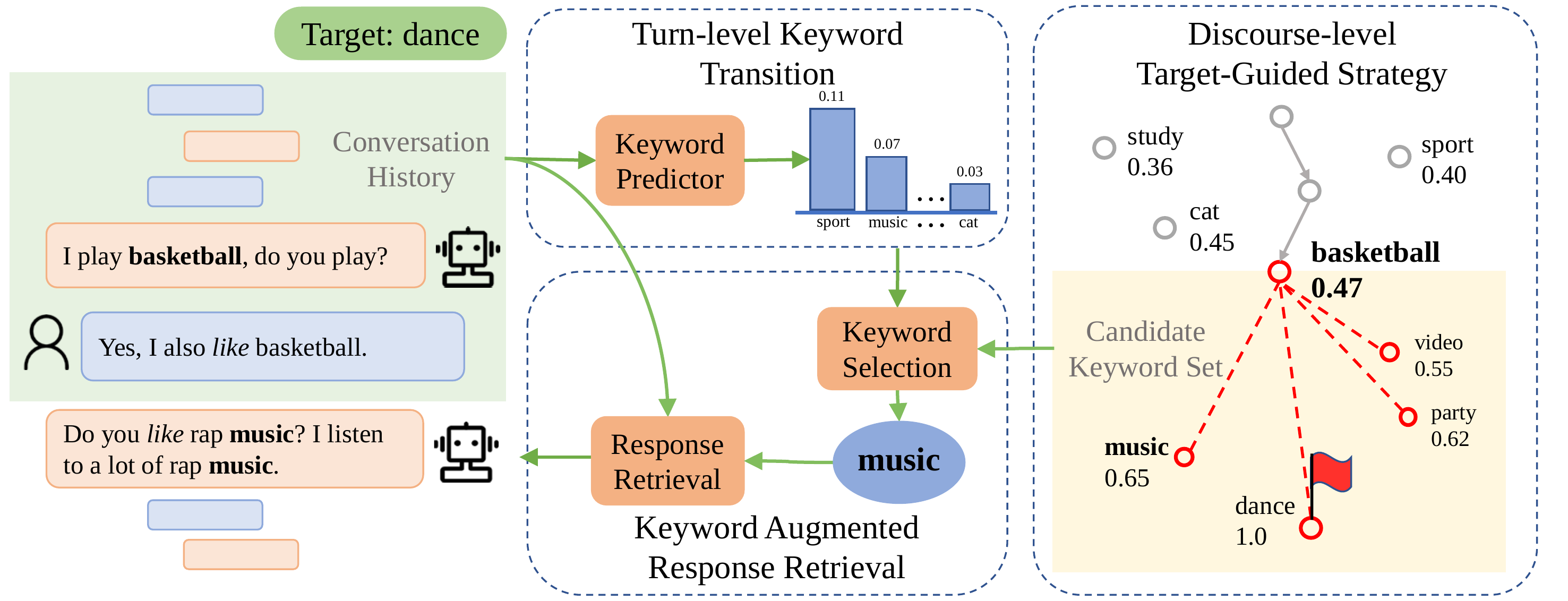}
\caption{
Solution Overview. The left panel shows an on-going conversation with a designated target \emph{dance}.
The discourse-level target-guided module (right panel, section~\ref{sec:model-disource}) first picks a set of valid candidate keywords for the next system response. The turn-level keyword transition module (middle panel, section~\ref{sec:model-turn}) computes a distribution over candidate keywords. The most likely valid keyword (\emph{music}) is then selected, and fed into the keyword-augmented response retrieval module (middle panel, section~\ref{sec:model-retrieval}) for producing the next response.
}
\label{fig:model}
\end{figure*}

\section{The Proposed Approach}\label{sec:method}
We explore a solution that addresses the two desiderata separately. In particular, we maintain smooth conversation transition by turn-level supervised learning on open-domain chat data, and we inject target-guiding behavior with a rule-based guiding strategy. Further, to enable effective control over the transition and guiding strategy, we decouple the decision-making process and utterance generation by explicitly modeling the intended coarse-grained keywords in the next system utterance.

Thus the system consists of several core modules, including a turn-level keyword transition predictor (section~\ref{sec:model-turn}), a discourse-level target-guiding strategy (section~\ref{sec:model-disource}), and a response retriever (section~\ref{sec:model-retrieval}). 



\subsection{Turn-level Keyword Transition}\label{sec:model-turn}
Given the conversation history at each dialogue turn, this module aims to predict keywords of the next response that is appropriate in the conversation context. 
This part is agnostic to the end target, and therefore aligns with the conventional chit-chat objective. We thus can use any open-ended chat data with extracted utterance keywords to learn the prediction module in a supervised manner. We present such a dataset that we posit is particularly suitable for the learning in section~\ref{sec:data}.

Architecturally, we study three different approaches as representative paradigms for predicting the next-turn keyword distribution, including pairwise keyword linear transition, neural-based prediction, and kernel-based method. 



\paragraph{Pairwise PMI-based Transition}
The most straightforward way for keyword transition is to construct a keyword pairwise matrix that characterizes the association between keywords in the observed conversation data. We use pointwise mutual information (PMI)~\citep{church1990word} as the measure, which, given two keywords $w_i$ and $w_j$, computes likeliness of $w_j\to w_i$ with 
\begin{equation}
\small
\text{PMI}(w_i, w_j)=\log p(w_i|w_j) / p(w_i),
\end{equation}
where $p(w_i|w_j)$ is the ratio of transitioning to $w_i$ in the next turn given $w_j$ in the current turn, and $p(w_i)$ is the ratio of $w_i$ occurrence. Both quantities can be directly counted from the conversation data beforehand. At test time, we
first use a keyword extractor (section~\ref{sec:data}) to extract keywords of the current utterance. Assuming all these keywords are independent, for each candidate next keyword, we sum up their PMI scores w.r.t the candidate. The resulting candidate scores are then normalized to obtain a distribution over keywords in the next turn.

The approach enjoys simplicity and interpretability, yet can suffer from data sparsity and perform poorly with {\it a priori} unseen transition pairs.



\paragraph{Neural-based Prediction}
The second approach predicts the next keywords with a neural network in an end-to-end manner. More concretely, we first use a recurrent network to encode the conversation history, and feed the resulting features to a prediction layer to obtain a distribution over keywords for the next turn.
The network is learned by maximizing the likelihood of observed keywords in the data. The neural approach is straightforward, but can rely on a large amount of data for learning.


\paragraph{Hybrid Kernel-based Method}
We further study a hybrid approach that combines neural feature extraction with pairwise closeness measuring. Specifically, given a pair of a current keyword and a candidate next keyword, we follow~\citep{xiong2017end} by first measuring the cosine similarity of their normalized word embeddings, and feeding the quantity to a kernel layer consisting of $K$ RBF kernels. The output of the kernel layer is a $K$-dimension kernel feature vector, which is then fed to a single-unit dense layer for a candidate score. The score is finally normalized across all candidate keywords to yield the candidate probability distribution. 
If the current turn has multiple keywords, the corresponding multiple $K$-dimension kernel features are first summed up before feeding to the dense layer. Thus, the intermediate kernel layer serves as a soft aggregation mechanism to account for multiple-to-one keyword transition. 
The parameters are learned in the same way as in the neural-based prediction method. Our empirical study shows the hybrid approach provides the strongest performance.

\subsection{Discourse-level Target-Guided Strategy}\label{sec:model-disource}
This module aims to fulfill the end target by proactively driving the discussion topic forward in the course of the conversation. As noted above, there is typically no data available for direct learning of such a strategy. Fortunately, the augmentation of interpretable coarse-grained keywords enables us to apply a simple yet effective rule to this end.

We constrain that the keyword of each turn must move strictly closer to the end target compared to those of preceding turns. Figure~\ref{fig:model}, right part, illustrates the rule at a particular step. Given the keyword \emph{Basketball} of the current turn and its closeness score (0.47) to the target \emph{Dance}, the only valid candidate keywords for the next turn are those with higher target closeness, such as \emph{Party} with a closeness score of 0.62. On the other hand, transitioning from \emph{Basketball} to \emph{Sport} is not allowed in the context as it does not move towards the target. More concretely, we use cosine similarity between normalized word embeddings as the measure of keyword closeness. 

At each turn for predicting the next keyword, the above constraint first collects a set of valid candidates, and the turn-level transition module samples or picks the most likely one the from the set according to the keyword distribution. In this way, the predicted keyword for next response can be both a smooth transition and an effective step towards the target.

\subsection{Keyword-augmented Response Retrieval}\label{sec:model-retrieval}
The final module in the system aims to produce a response conditioning on both the conversation history and the predicted keyword. In this work, we use a retrieval-based approach, though a generation-based method can also be readily plugged in.

The architecture of the module is adapted from the previous work~\cite{wu2016sequential} with augmented keyword conditioning. More concretely, we use recurrent networks to encode the input conversation history and keyword, as well as each of the candidate responses from a database (e.g., all utterances in the training set). We then compute the element-wise product between the candidate feature with the history feature, and between the candidate feature with the keyword feature, respectively. The resulting two vectors are concatenated and fed to a final single-unit dense layer with sigmoid to get the matching probability of the candidate response.

Same as the turn-level transition module, the conditional response retrieval module can also be learned with open-ended conversation data in a supervised manner. That is, we maximize the likelihood of observed response given its history and predicted keyword, while minimizing the likelihood of randomly sampled negative responses. Section~\ref{sec:data} presents more details of the data and negative responses.

\begin{table}[t]
\small
\centering
\begin{tabular}{r l l l} 
\cmidrule[\heavyrulewidth]{1-4}
& {\bf Train} & {\bf Val} & {\bf Test} \\ \cmidrule{1-4}
  \#Conversations & 8,939 & 500 & 500\\
  \#Utterances & 101,935 & 5,602 & 5,317\\
  \#Keyword types & 2,678 & 2,080 & 1,571\\
  \#Avg. keywords & 2.1 & 2,1 & 1.9 \\
\cmidrule[\heavyrulewidth]{1-4}
\end{tabular}
\caption{Data Statistics. The last row is the average number of keywords in each utterance. The vocabulary size is around 19K.}

\label{table:data}
\end{table}
\begin{table}[t]
\centering
\small
\begin{tabular}{@{}R{0.3cm} L{6.6cm}@{}}
\cmidrule[\heavyrulewidth]{1-2}
{\bf A:} & Hi ! I am from \underline{India} . where are you from?\\[4pt] 
{\bf B:} & I'm from \underline{Portland}. I just got back from a long \underline{walk}.\\[4pt] 
{\bf A:} & I just got back from \underline{coaching} \underline{swimming} at the \underline{pool}. Walking where ?\\[4pt]
{\bf B:} & I like to walk in \underline{parks} for good \underline{health}. No soft \underline{drinks} for me either! \\
... & ... \\ [4pt] 
\cmidrule[\heavyrulewidth]{1-2}
\end{tabular}
\vspace{-6pt}
\caption{
An Example Conversation. Only the first 4 utterances are shown. Keywords of each utterance are marked with underline.
}
\label{table:example}
\end{table}

\section{Dataset}\label{sec:data}
We next describe a large conversation dataset that can be useful for studying the task and has been used in our solution. The dataset is derived from the PersonaChat corpus~\citep{zhang2018personalizing} where crowdworkers were asked to chat naturally with given persona. The conversations cover a broad range of topics such as work, family, personal interest, etc; and the discussion topics change frequently during the course of the conversations. These properties make the conversations particularly suitable for learning smooth, natural transitions \emph{at each turn}. Note that, however, the conversations do not necessarily suit for learning discourse-level strategies, as they were originally created without end targets and do not exhibit target-guided behaviors.

To adapt the corpus for turn-level keyword transition in our new setting, we obtain all conversations while discarding the associated persona information. We then augment the data by automatically extracting keywords of each utterance. Specifically, we apply a rule-based keyword extractor which combines TF-IDF and Part-Of-Speech features for scoring word salience. More details are provided in supplementary materials. We re-split the data into train/valid/test sets, where the test set contains 500 conversations with relatively frequent keywords. Table~\ref{table:data} lists the data statistics. An example conversation with the extracted keywords is shown in Table~\ref{table:example}.

The resulting dataset is used in our solution for training both the turn-level transition module (section~\ref{sec:model-turn}) and the response retrieval module (section~\ref{sec:model-retrieval}). We follow the retrieval-based chit-chat literature~\citep{wu2016sequential} and randomly sample 19 negative responses for each turn as the negative responses for training.

\begin{table*}[t]
\small
\centering
\begin{tabular}{r|lllll|llll}
\cmidrule[\heavyrulewidth]{1-10}
  & \multicolumn{5}{c|}{\textbf{Keyword Prediction}} & \multicolumn{4}{c}{\textbf{Response Retrieval}}\\[4pt]
  \textbf{System} & $\bm{R_{w}@1}$ &  $\bm{R_{w}@3}$ &  $\bm{R_{w}@5}$ & $\bm{P@1}$ &  \textbf{Cor.} & $\bm{R_{20}@1}$ & $\bm{R_{20}@3}$ &$\bm{R_{20}@5}$ & \textbf{MRR}\\
  \cmidrule[\heavyrulewidth]{1-10}
  Retrieval & - & - & - & - & - & 0.5196 & 0.7636 & 0.8622 & 0.6661\\
  Ours-Random & 0.0005 & 0.0015 & 0.0025 & 0.0009 & 0.4995 & 0.5187 & 0.7619 & 0.8631 & 0.6650\\
  Ours-PMI & 0.0585 & 0.1351 & 0.1872 & 0.0871 & 0.7974 & 0.5441 & \textbf{0.7839} & 0.8716 & 0.6847\\
  Ours-Neural & 0.0609 & 0.1324 & 0.1825 & 0.1006 & 0.8075  & 0.5395 & 0.7801 & 0.8790 & 0.6816\\
  Ours-Kernel & \textbf{0.0642} & \textbf{0.1431} & \textbf{0.1928} & \textbf{0.1191} & \textbf{0.8164} & \textbf{0.5486} & 0.7827 & \textbf{0.8845} & \textbf{0.6914}\\
\cmidrule[\heavyrulewidth]{1-10}
\end{tabular}
\vspace{-6pt}
\caption{Results of Turn-level Evaluation.}
\label{table:turn-level}
\vspace{-10pt}
\end{table*}

\section{Experiments}

\subsection{Experimental Setup}

\paragraph{Baselines and Comparison Systems}

We evaluate a diverse set of approaches for comparison and ablation study.

\textbf{Retrieval}~\citep{wu2016sequential} is the conventional retrieval-based chitchat system which does not permit an end target and is not augmented with coarse-grained utterance keywords. The system thus cannot be deployed for target-guided conversation, and is used to provide reference performance in terms of different metrics in the experiments. The model architecture is adapted from the prior work, the same as used in our full system except for the keyword conditioning part.


\textbf{Retrieval-Stgy} augments the above base retrieval system with the proposed target-guided strategy (section~\ref{sec:model-disource}). Specifically, it first extracts the keywords of current utterance with the extractor used in section~\ref{sec:data}, and applies the target-guided rule to obtain a set of candidate keywords. The base retrieval model is then used to retrieve a response containing at least one keyword from the keyword set. Such a pipeline approach achieves a strong baseline performance, as shown in the following.

\textbf{Ours } As in section~\ref{sec:model-turn}, our full system has several variants in the turn-level keyword transition module, including the \textbf{PMI}, \textbf{Neural}, and \textbf{Kernel} methods. For comparison, we also use a \textbf{Random} method which randomly picks a keyword for next response.

\paragraph{Training Details}
We use the same configuration for the common parts of all agents. We apply a single-layer GRU~\cite{chung2014empirical} in all encoders. Both the word embedding and hidden dimensions are set to 200. 
We use GloVe~\cite{pennington2014glove} to initialize word embeddings.
We apply Adam optimization~\cite{kingma2014adam} with an initial learning rate of $0.001$ and annealing to $0.0001$ in 10 epochs. Systems are implemented with a text generation toolkit Texar~\citep{hu2019texar}.

\subsection{Turn-level Evaluation}

We first evaluate the performance of each conversation turn, in terms of both turn-level keyword prediction and response selection.
That is, we disable the discourse-level target constraint, and focus on measuring how accurate the systems can predict the next keyword and retrieve the correct response on the test set of the conversation data. The evaluation largely follows the protocol of previous chit-chat systems~\citep[e.g.,][]{wu2016sequential}, and validates the effect of the keyword-augmented conversation production.

\paragraph{Evaluation metrics} 
For the keyword prediction task, we measure three metrics: (1) $\bm{R_{w}@K}$: keywords recall at position $K$ ($=1,3,5$) in all (over 2600) possible keywords, (2) $\bm{P@1}$: precision at the first position, and (3) {\bf Cor.}: the word embedding based correlation score~\cite{liu2016not}. 

For the response selection task, we randomly sample 19 negative responses for each test case, and calculate $\bm{R_{20}@K}$, i.e., recall at position $K$ in the 20 candidate (positive and negative) responses, as well as {\bf MRR}, the mean reciprocal rank.

\paragraph{Results} 
Table~\ref{table:turn-level} shows the evaluation results. Our system with Kernel transition module outperforms all other systems in terms of all metrics on both two tasks, expect for $R_{20}@3$ where the system with PMI transition performs best. The Kernel approach can predict the next keywords more precisely. In the task of response selection, our systems that are augmented with predicted keywords significantly outperform the base Retrieval approach, showing predicted keywords are helpful for better retrieving responses by capturing coarse-grained information of the next utterances. Interestingly, the
system with Random transition has a close performance to the base Retrieval model, indicating that the erroneous keywords can be ignored by the system after training.

\subsection{Target-guided Conversation Evaluation}
We next evaluate system performance in the proposed target-guided conversation setup, with both automatic simulation-based evaluation and human evaluation.

\subsubsection{Self-Play Simulation}

\begin{table}[t]
\small
\centering
\begin{tabular}{r l l} 
\cmidrule[\heavyrulewidth]{1-3}
  \textbf{System} & \textbf{Succ.~(\%)} & \textbf{\#Turns}  \\ \cmidrule{1-3}
  Retrieval & 9.8 & 3.26  \\
  Retrieval-Stgy & 67.2 & 6.56 \\
  Ours-PMI & 47.4 & 5.12 \\
  Ours-Neural & 51.6 & 4.29 \\
  Ours-Kernel & \textbf{75.0} & 4.20 \\
\cmidrule[\heavyrulewidth]{1-3}
\end{tabular}
\vspace{-10pt}
\caption{Results of Self-Play Evaluation.}
\label{table:auto}
\end{table}

Following the experimental settings in prior work~\cite{lewis2017deal,li2016deep}, we developed a task simulator to automatically produce target-guided conversations. Specifically, we use the base Retrieval agent to play the role of human which retrieves a response without knowing the end target. 
The simulator randomly picks a keyword as the end target, and an utterance as the starting point. Each agent then chats with the Retrieval system, trying to guide the conversation to the given target. To automatically evaluate whether the target is achieved, we use WordNet~\cite{miller1998wordnet} to identify keywords that are semantically close to the end target. More concretely, if a keyword in an utterance (by either the agent under test or Retrieval) has a WordNet information content similarity score higher than 0.9, we consider the target is successfully achieved. To avoid infinite conversation without ever reaching the target,
we set a maximum allowed number of turns, which is 8 in our experiment. That is, an agent that does not achieve the target after producing 8 responses is considered to fail in the case.
We measure the success rate of achieving the targets (\textbf{Succ.}) and the average number of turns used to reach a target (\textbf{\#Turns}). 

Table~\ref{table:auto} shows the results of 500 simulations for each of the comparison systems. Our system with Kernel transition obtains the highest success rate, significantly improving over other approaches.
The success rate of the base Retrieval agent is lower than 10\%, which proves that a chitchat agent without a target-guided strategy can hardly accomplish our task.
The Retrieval-Stgy agent has a relatively high success rate, while taking more turns (6.56) to accomplish this. This is partially due to the lack of coarse-grained keyword modeling and transition.
We further note that, in the Kernel system, around 81\% of predicted keywords eventually occur in the produced utterances, indicating that the predicted keywords have a great impact on the retrieval module.

\begin{table}[t]
\small
\centering
\begin{tabular}{r l l} 
\cmidrule[\heavyrulewidth]{1-3}
  \textbf{System} & \textbf{Succ.~(\%)} & \bf Smoothness  \\ \cmidrule{1-3}
  Retrieval & 18 & 3.26  \\
  Retrieval-Stgy & 66 & 3.24 \\
  Ours-PMI & 52 & 3.00  \\
  Ours-Neural &  56 & 2.94  \\
  Ours-Kernel &  \textbf{76} & \textbf{3.40} \\
\cmidrule[\heavyrulewidth]{1-3}
\end{tabular}
\vspace{-10pt}
\caption{Results of the Human Rating.}
\label{table:human1}
\end{table}

\begin{table}[t]
\centering
\small
\begin{tabular}{@{}r l l l@{}}
\cmidrule[\heavyrulewidth]{1-4}
 & \multirow{2}{1.4cm}{{\bf Kernel Better(\%)}}  &\multirow{2}{1.5cm}{{\bf No Prefer(\%)}} & \multirow{2}{1.5cm}{{\bf Kernel Worse(\%)}} \\
 & & &\\
 \cmidrule{1-4}
Retrieval-Stgy  & {\bf 34} & 40 & 26\\
Ours-PMI & {\bf 52} & 22 & 26 \\
Ours-Neural & {\bf 56} & 28 & 16 \\
\cmidrule[\heavyrulewidth]{1-4}
\end{tabular}
\vspace{-6pt}
\caption{Results of Human Ranking.}
\label{table:human2}
\end{table}

\begin{table*}[t]
\centering
\small
\begin{tabular}{l|l}
\cmidrule[\heavyrulewidth]{1-2}
\textbf{Agent: Retrieval-Stgy\quad Target: \red{\underline{Dance}}} &
\textbf{Agent: Ours-Kernel\quad  Target: \red{\underline{Dance}}}\\
\cmidrule{1-2}
 \begin{tabular}{@{}R{0.2cm} L{7cm}@{}}
\textbf{A:}& Hi, how are you today?\\
\textbf{H:}& Fine. Just finished \blue{\emph{riding}} in the countryside.\\
\textbf{A:}& That sounds \red{\emph{relaxing}}. I love the outdoors.\\
\textbf{H:}& I rode along a long black \blue{\emph{river}}.\\
\textbf{A:}& Nice , do you like the \red{\emph{lake}}? I do I go often.\\
\textbf{H:}& Yes. I like \blue{\emph{swimming}} in the lake.\\
\textbf{A:}& I've never been surfing but used to do a lot of traveling across the us playing \red{\emph{guitar}}.\\
\textbf{H:}& Nice. I also like playing guitar and other \blue{\emph{musical instruments}}.\\
\textbf{A:}& Oh that's nice, I'm more into classical \red{\emph{music}}.\\
\textbf{H:}& I am interested in \blue{\emph{pop music}}.\\
\textbf{A:}& Oh cool! I really like music and \red{\emph{singing}}.\\
\textbf{H:}& I enjoy \blue{\emph{singing}}. Although I cannot sing well.\\
\textbf{A:}& Music is life. \red{\underline{\emph{Dancing}}} to anything with a beat is the best.\\
 \end{tabular}
&
 \begin{tabular}{@{}R{0.2cm} L{7cm}@{}}
\textbf{A:}& Hi, how are you today?\\
\textbf{H:}& Fine. Just finished \blue{\emph{riding}} in the countryside.\\
\textbf{A:}& I love riding my bike. It is great exercise and fun. \red{(\emph{bike})}\\
\textbf{H:}& I rode along a long black \blue{\emph{river}}.\\
\textbf{A:}& That sounds very exciting. You can ride bikes, listen to musics there too. \red{(\emph{music})}\\
\textbf{H:}& That's great. My favorite \blue{\emph{singer}} is Taylor Swift.\\
\textbf{A:}& Yes I love to sing! I can sing a lot of songs, can you? \red{(\emph{sing})}\\
\textbf{H:}& Yes, although I cannot \blue{\emph{sing}} well.\\
\textbf{A:}& Dancing? I love \red{\underline{\emph{dancing}}}! Singing too!
\\
\\
\\
\\
\\
 \end{tabular} 
 \\

\cmidrule{1-2}\morecmidrules\cmidrule{1-2}
\end{tabular}
\begin{tabular}{l|l}
\textbf{Agent: Retrieval-Stgy\quad  Target: \red{\underline{McDonald's}}} &
\textbf{Agent: Ours-Kernel\quad  Target: \red{\underline{McDonald's}}}\\
\cmidrule{1-2}
 \begin{tabular}{@{}R{0.2cm} L{7cm}@{}}
\textbf{A:}& Hello, how are you doing?\\
\textbf{H:}& Hi, I am great just finishing up my \blue{\emph{homework}}. How are you?\\
\textbf{A:}& Oh nothing much just \red{\emph{reading}}.\\
\textbf{H:}& Which kind of \blue{\emph{books}} are you reading?\\
\textbf{A:}& Reading comic \red{\emph{books}}, I love them. What about you?\\
\textbf{H:}& I enjoy reading sci-fi and romantic \blue{\emph{stories}}.\\
\textbf{A:}& That sounds helpful for \red{\emph{writing}}.\\
\textbf{H:}& Yeah, but I do not good at writing \blue{\emph{articles}}\\
\textbf{A:}& Those were good books, what do you like to \red{\emph{eat}}?\\
\textbf{H:}& \blue{\emph{Vegetables}} and fruit salad are my favorite.\\
\textbf{A:}& That reminds me of spicy \red{\emph{food}}.\\
\textbf{H:}& Oh spicy food is also very \blue{\emph{delicious}}.\\
\textbf{A:}& Nothing better than bacon for \red{\emph{breakfast}}.\\
\textbf{H:}& I often eating \blue{\emph{eggs}} for breakfast.\\
\textbf{A:}& I love cheese \red{\emph{tacos}} too.\\
\textbf{H:}& It's too late now, and I'll ride my bike to \blue{\emph{home}}.\\
 \end{tabular}
& 
 \begin{tabular}{@{}R{0.2cm} L{7cm}@{}}
\textbf{A:}& Hello, how are you doing?\\
\textbf{H:}& Hi, I am great just finishing up my \blue{\emph{homework}}. How are you?\\
\textbf{A:}& Just got done eating some chicken and rice after I got done with my gym session. \red{(\emph{eat})}\\
\textbf{H:}& I am gonna eat some vegetable for \blue{\emph{dinner}}.\\
\textbf{A:}& So you are vegetarian? I am from los angeles so there are lots of vegetarian restaurants. \red{(\emph{restaurant})} \\
\textbf{H:}& I am not a vegetarian. Fried \blue{\emph{chicken}} is  my favorite.\\
\textbf{A:}& Oh boy, when I go to \red{\underline{\emph{McDonald's}}}, I always get fries for my 2 dogs.\\
\\
\\
\\
\\
\\
\\
 \end{tabular}
 \\
\cmidrule[\heavyrulewidth]{1-2}
\end{tabular}

\vspace{-10pt}
\caption{Example conversations between human (\textbf{H}) and two different agents (\textbf{A}), with the same targets and starting utterances. 
Keywords selected or predicted by the agents are highlighted in {\color{red} red} and \emph{italic}, and keywords mentioned by human are highlighted in {\color{blue} blue} and \emph{italic}. As keywords predicted by the Kernel agent do not necessarily occur in the retrieved utterances, we put them to the end of each sentence. Targets achieved at the end of conversations are underlined.
We present the examples in case-sensitive format for readability. All tokens are in lowercase in the program.}
\label{table:qual}
\end{table*}

\subsubsection{Human Evaluation}
We finally perform human evaluation for a more thorough system comparison in terms of different aspects. Specifically, we use the DialCrowd toolkit~\cite{lee2018dialcrowd} to setup human evaluation interfaces, and undertook two types of human studies as below. 

The first evaluation is to measure the system performance in terms of the two key desiderata, namely target achievement and transition smoothness, respectively. We first build 50 test cases, each of which has a target and a starting utterance. In each test case, a human turker is asked to converse with a randomly selected agent. The agent informs the turker when it thought the target is achieved or has reached the maximum number of turns (which is set to 8). Then the turker is presented with the designated target, and is asked to judge whether the target has been achieved, as well as rate transition smoothness during the conversation with a score ranging from 1 (strongly bad) to 5 (strongly good). All agents are evaluated on all test cases.

Table~\ref{table:human1} shows the results of the first evaluation. Our Kernel agent clearly outperforms all other comparison systems in terms of both success rate and transition smoothness. Note that the success rate results of all agent are consistent with those in simulation (Table~\ref{table:auto}). Comparing the base Retrieval agent and the augmented Retrieval-Stgy agent, we can see that Retrieval-Stgy has almost the same smoothness with Retrieval but achieves a much higher success rate. This validates that our discourse-level strategy (section~\ref{sec:model-disource}) is indeed effective for target-guided conversations.

The second evaluation compares our best-performing Kernel agents with other agents side-by-side. Specifically, we ask a human turker to converse with the Kernel agent and a randomly selected comparison agent in the same test case. We then ask the turker to rank the two conversations by considering all the criteria. Turkers can also choose ``no preference'' if the conversations are equally good or bad. To avoid any bias, in each test case, we randomly pick one from the pair of agents to converse first, and we let the turker to decide when to stop to avoid revealing the target too early. As above, we evaluate on 50 test cases for each pair of agents.

Table~\ref{table:human2} shows the results of the second evaluation. We see that our Kernel system consistently outperforms the comparison methods with 30-50\% wins.

\begin{table}[t]
\centering
\small
\begin{tabular}{R{0.2cm} L{6.8cm}@{}}
\cmidrule[\heavyrulewidth]{1-2}

 & \textbf{Agent: Ours-Kernel\quad Target: \red{\underline{Listen}}} \\

\cmidrule{1-2}
\cmidrule{1-2}
{\bf A:} & Hi! How are you? I am very exciting about my new job!\\
{\bf H:} & Fine. What is your new \blue{\emph{job}}? \\
{\bf A:} & That sounds exciting. \red{(\emph{sound})}\\
{\bf H:} & I like reading \blue{\emph{books}}.\\
{\bf A:} & What kind of books do you read? \red{(\emph{book})}\\
{\bf H:} & I read \blue{\emph{comics}}.\\
{\bf A:} & I do not have TV. \red{(\emph{TV})}\\
{\bf H:} & Watching \blue{\emph{animation}} is fun.\\
{\bf A:} & What position d you play? \red{(\emph{play})}\\
{\bf H:} & I play a lot of \blue{\emph{sports}}.\\
{\bf A:} & Oh really? What kind of music? \red{\underline{\emph{Listen}}} to classical. \\

\cmidrule[\heavyrulewidth]{1-2}
\end{tabular}
\vspace{-6pt}
\caption{An erroneous example.}
\label{table:fail}
\end{table}

\subsection{Qualitative Study}
We take a close look at the model performance by studying the conversation examples from different agents in human evaluation.

Table~\ref{table:qual} shows the conversations between human and agents given targets \emph{dance} and \emph{McDonald's}, respectively. We can see that, in general, our Kernel agent can accomplish the task in fewer turns than the Retrieval-Stg agent. In the first case, the Kernel agent guides the conversation from \emph{ride} to the crucial topic \emph{music} smoothly and quickly, and then achieves the target word \emph{dance} naturally. In contrast, the Retrieve-Stgy agent is trapped in open-ended chats for the first three turns and does not reach the target until the 7th turn. In the second case, the target \emph{McDonald's} is relatively uncommon in our dataset. The kernel agent succeeded to achieve the target in the 4th turn while the Retrieval-Stgy agent failed to reach the target within the maximally allowed number of turns.

Table~\ref{table:fail} shows a failure case by our Kernel agent. Although the agent successfully achieved the target, it sometimes makes non-smooth keyword transition without a clear logic. For instance, the final utterance of the agent, though reaching the target \emph{listen}, is not appropriate in the conversation context (e.g., in the presence of human's preceding keyword \emph{sports}).

\section{Conclusions \& Discussions}
We have studied the problem of target-guided open-domain conversation, where an agent converses naturally with the human and proactively guides the conversation to a designated end target. We propose a modular solution with coarse-grained keywords as a logical backbone, and use partial supervision and heuristic rules to achieve the task. We also derive a dataset for the study. Quantitative and human evaluations demonstrate promising and improved results of our approach.

This work presents an initial attempt to bridge the gap between open-domain chit-chat and task-oriented dialogue. A target-guided agent can be deployed in practice to converse with users engagingly and guide the users to trigger task-oriented systems (e.g., reserving a restaurant) in the end. An open-domain agent with control over the conversation strategy and end target can also be useful in education, psychotherapy, and others as discussed in section~\ref{sec:intro}. Our treatment of utterance action and conversation target through simple keywords can be preliminary in terms of complex real applications. It would be exciting to explore more sophisticated modeling to enable more fine-grained control on both sentence~\citep{hu2017toward} and discourse levels~\citep{williams2017hybrid,fang2018sounding}.

\balance
\bibliography{acl2019}

\begin{thebibliography}{42}
\expandafter\ifx\csname natexlab\endcsname\relax\def\natexlab#1{#1}\fi

\bibitem[{Cao et~al.(2018)Cao, Lazaridou, Lanctot, Leibo, Tuyls, and
  Clark}]{cao2018emergent}
Kris Cao, Angeliki Lazaridou, Marc Lanctot, Joel~Z Leibo, Karl Tuyls, and
  Stephen Clark. 2018.
\newblock Emergent communication through negotiation.
\newblock In \emph{ICLR}.

\bibitem[{Chung et~al.(2014)Chung, Gulcehre, Cho, and
  Bengio}]{chung2014empirical}
Junyoung Chung, Caglar Gulcehre, KyungHyun Cho, and Yoshua Bengio. 2014.
\newblock Empirical evaluation of gated recurrent neural networks on sequence
  modeling.
\newblock \emph{arXiv preprint arXiv:1412.3555}.

\bibitem[{Church and Hanks(1990)}]{church1990word}
Kenneth~Ward Church and Patrick Hanks. 1990.
\newblock Word association norms, mutual information, and lexicography.
\newblock \emph{Computational linguistics}, 16(1):22--29.

\bibitem[{DeVault et~al.(2015)DeVault, Mell, and Gratch}]{devault2015toward}
David DeVault, Johnathan Mell, and Jonathan Gratch. 2015.
\newblock Toward natural turn-taking in a virtual human negotiation agent.
\newblock In \emph{2015 AAAI Spring Symposium Series}.

\bibitem[{Dhingra et~al.(2017)Dhingra, Li, Li, Gao, Chen, Ahmed, and
  Deng}]{dhingra2017towards}
Bhuwan Dhingra, Lihong Li, Xiujun Li, Jianfeng Gao, Yun-Nung Chen, Faisal
  Ahmed, and Li~Deng. 2017.
\newblock Towards end-to-end reinforcement learning of dialogue agents for
  information access.
\newblock In \emph{ACL}.

\bibitem[{Fang et~al.(2018)Fang, Cheng, Sap, Clark, Holtzman, Choi, Smith, and
  Ostendorf}]{fang2018sounding}
Hao Fang, Hao Cheng, Maarten Sap, Elizabeth Clark, Ari Holtzman, Yejin Choi,
  Noah~A Smith, and Mari Ostendorf. 2018.
\newblock Sounding board: A user-centric and content-driven social chatbot.
\newblock In \emph{NAACL System Demonstrations}.

\bibitem[{Ghazvininejad et~al.(2018)Ghazvininejad, Brockett, Chang, Dolan, Gao,
  Yih, and Galley}]{ghazvininejad2018knowledge}
Marjan Ghazvininejad, Chris Brockett, Ming-Wei Chang, Bill Dolan, Jianfeng Gao,
  Wen-tau Yih, and Michel Galley. 2018.
\newblock A knowledge-grounded neural conversation model.
\newblock In \emph{AAAI}.

\bibitem[{He et~al.(2017)He, Balakrishnan, Eric, and Liang}]{he2017learning}
He~He, Anusha Balakrishnan, Mihail Eric, and Percy Liang. 2017.
\newblock Learning symmetric collaborative dialogue agents with dynamic
  knowledge graph embeddings.
\newblock In \emph{ACL}.

\bibitem[{He et~al.(2018)He, Chen, Balakrishnan, and Liang}]{he2018decoupling}
He~He, Derek Chen, Anusha Balakrishnan, and Percy Liang. 2018.
\newblock Decoupling strategy and generation in negotiation dialogues.
\newblock \emph{arXiv preprint arXiv:1808.09637}.

\bibitem[{Hu et~al.(2019)Hu, Shi, Tan, Wang, Yang, Zhao, He, Qin, Wang
  et~al.}]{hu2019texar}
Zhiting Hu, Haoran Shi, Bowen Tan, Wentao Wang, Zichao Yang, Tiancheng Zhao,
  Junxian He, Lianhui Qin, Di~Wang, et~al. 2019.
\newblock Texar: A modularized, versatile, and extensible toolkit for text
  generation.
\newblock In \emph{ACL System Demonstrations}.

\bibitem[{Hu et~al.(2017)Hu, Yang, Liang, Salakhutdinov, and
  Xing}]{hu2017toward}
Zhiting Hu, Zichao Yang, Xiaodan Liang, Ruslan Salakhutdinov, and Eric~P Xing.
  2017.
\newblock Toward controlled generation of text.
\newblock In \emph{ICML}.

\bibitem[{Kingma and Ba(2014)}]{kingma2014adam}
Diederik~P Kingma and Jimmy Ba. 2014.
\newblock Adam: A method for stochastic optimization.
\newblock \emph{arXiv preprint arXiv:1412.6980}.

\bibitem[{Lee et~al.(2018)Lee, Zhao, Black, and Eskenazi}]{lee2018dialcrowd}
Kyusong Lee, Tiancheng Zhao, Alan~W Black, and Maxine Eskenazi. 2018.
\newblock Dial{C}rowd: A toolkit for easy dialog system assessment.
\newblock In \emph{SIGDAIL}, pages 245--248.

\bibitem[{Lewis et~al.(2017)Lewis, Yarats, Dauphin, Parikh, and
  Batra}]{lewis2017deal}
Mike Lewis, Denis Yarats, Yann~N Dauphin, Devi Parikh, and Dhruv Batra. 2017.
\newblock Deal or no deal? end-to-end learning for negotiation dialogues.
\newblock \emph{arXiv preprint arXiv:1706.05125}.

\bibitem[{Li et~al.(2015)Li, Galley, Brockett, Gao, and
  Dolan}]{li2015diversity}
Jiwei Li, Michel Galley, Chris Brockett, Jianfeng Gao, and Bill Dolan. 2015.
\newblock A diversity-promoting objective function for neural conversation
  models.
\newblock \emph{arXiv preprint arXiv:1510.03055}.

\bibitem[{Li et~al.(2016{\natexlab{a}})Li, Galley, Brockett, Gao, and
  Dolan}]{li2016persona}
Jiwei Li, Michel Galley, Chris Brockett, Jianfeng Gao, and Bill Dolan.
  2016{\natexlab{a}}.
\newblock A persona-based neural conversation model.
\newblock \emph{arXiv preprint arXiv:1603.06155}.

\bibitem[{Li et~al.(2016{\natexlab{b}})Li, Monroe, Ritter, Galley, Gao, and
  Jurafsky}]{li2016deep}
Jiwei Li, Will Monroe, Alan Ritter, Michel Galley, Jianfeng Gao, and Dan
  Jurafsky. 2016{\natexlab{b}}.
\newblock Deep reinforcement learning for dialogue generation.
\newblock \emph{arXiv preprint arXiv:1606.01541}.

\bibitem[{Li et~al.(2018)Li, Kahou, Schulz, Michalski, Charlin, and
  Pal}]{li2018towards}
Raymond Li, Samira~Ebrahimi Kahou, Hannes Schulz, Vincent Michalski, Laurent
  Charlin, and Chris Pal. 2018.
\newblock Towards deep conversational recommendations.
\newblock In \emph{NeurIPS}, pages 9748--9758.

\bibitem[{Lin et~al.(2019)Lin, Wang, Yang, Shi, Xu, Liang, Xing, and
  Hu}]{wang2019toward}
Shuai Lin, Wentao Wang, Zichao Yang, Haoran Shi, Frank Xu, Xiaodan Liang, Eric
  Xing, and Zhiting Hu. 2019.
\newblock Toward unsupervised text content manipulation.
\newblock \emph{arXiv preprint arXiv:1901.09501}.

\bibitem[{Liu et~al.(2016)Liu, Lowe, Serban, Noseworthy, Charlin, and
  Pineau}]{liu2016not}
Chia-Wei Liu, Ryan Lowe, Iulian~V Serban, Michael Noseworthy, Laurent Charlin,
  and Joelle Pineau. 2016.
\newblock How not to evaluate your dialogue system: An empirical study of
  unsupervised evaluation metrics for dialogue response generation.
\newblock \emph{arXiv preprint arXiv:1603.08023}.

\bibitem[{Liu et~al.(2018)Liu, Chen, Ren, Feng, Liu, and
  Yin}]{liu2018knowledge}
Shuman Liu, Hongshen Chen, Zhaochun Ren, Yang Feng, Qun Liu, and Dawei Yin.
  2018.
\newblock Knowledge diffusion for neural dialogue generation.
\newblock In \emph{ACL}, volume~1, pages 1489--1498.

\bibitem[{Miller(1998)}]{miller1998wordnet}
George Miller. 1998.
\newblock \emph{Word{N}et: An electronic lexical database}.
\newblock MIT press.

\bibitem[{Pennington et~al.(2014)Pennington, Socher, and
  Manning}]{pennington2014glove}
Jeffrey Pennington, Richard Socher, and Christopher Manning. 2014.
\newblock Glove: Global vectors for word representation.
\newblock In \emph{EMNLP}.

\bibitem[{Ram et~al.(2018)Ram, Prasad, Khatri, Venkatesh, Gabriel, Liu, Nunn,
  Hedayatnia, Cheng, Nagar et~al.}]{ram2018conversational}
Ashwin Ram, Rohit Prasad, Chandra Khatri, Anu Venkatesh, Raefer Gabriel, Qing
  Liu, Jeff Nunn, Behnam Hedayatnia, Ming Cheng, Ashish Nagar, et~al. 2018.
\newblock Conversational {AI}: The science behind the {Alexa} prize.
\newblock \emph{arXiv preprint arXiv:1801.03604}.

\bibitem[{Raux et~al.(2005)Raux, Langner, Bohus, Black, and
  Eskenazi}]{raux2005let}
Antoine Raux, Brian Langner, Dan Bohus, Alan~W Black, and Maxine Eskenazi.
  2005.
\newblock Let's go public! taking a spoken dialog system to the real world.
\newblock In \emph{Ninth European conference on speech communication and
  technology}.

\bibitem[{Serban et~al.(2016)Serban, Sordoni, Lowe, Charlin, Pineau, Courville,
  and Bengio}]{serban2016hierarchical}
Iulian~Vlad Serban, Alessandro Sordoni, Ryan Lowe, Laurent Charlin, Joelle
  Pineau, Aaron Courville, and Yoshua Bengio. 2016.
\newblock A hierarchical latent variable encoder-decoder model for generating
  dialogues.
\newblock \emph{arXiv preprint arXiv:1605.06069}.

\bibitem[{Shang et~al.(2015)Shang, Lu, and Li}]{shang2015neural}
Lifeng Shang, Zhengdong Lu, and Hang Li. 2015.
\newblock Neural responding machine for short-text conversation.
\newblock \emph{arXiv preprint arXiv:1503.02364}.

\bibitem[{Shen et~al.(2017)Shen, Lei, Barzilay, and Jaakkola}]{shen2017style}
Tianxiao Shen, Tao Lei, Regina Barzilay, and Tommi Jaakkola. 2017.
\newblock Style transfer from non-parallel text by cross-alignment.
\newblock In \emph{NeurIPS}.

\bibitem[{Sordoni et~al.(2015)Sordoni, Galley, Auli, Brockett, Ji, Mitchell,
  Nie, Gao, and Dolan}]{sordoni2015neural}
Alessandro Sordoni, Michel Galley, Michael Auli, Chris Brockett, Yangfeng Ji,
  Margaret Mitchell, Jian-Yun Nie, Jianfeng Gao, and Bill Dolan. 2015.
\newblock A neural network approach to context-sensitive generation of
  conversational responses.
\newblock \emph{arXiv preprint arXiv:1506.06714}.

\bibitem[{Weizenbaum et~al.(1966)}]{weizenbaum1966eliza}
Joseph Weizenbaum et~al. 1966.
\newblock {ELIZA}---a computer program for the study of natural language
  communication between man and machine.
\newblock \emph{Communications of the ACM}, 9(1):36--45.

\bibitem[{Williams et~al.(2017)Williams, Asadi, and Zweig}]{williams2017hybrid}
Jason~D Williams, Kavosh Asadi, and Geoffrey Zweig. 2017.
\newblock Hybrid code networks: practical and efficient end-to-end dialog
  control with supervised and reinforcement learning.
\newblock In \emph{ACL}.

\bibitem[{Wu et~al.(2018)Wu, Liu, Chen, Zhao, Dong, Yu, Zhou, and
  Li}]{Wu2018MultiTurnRS}
Hua Wu, Yi~Liu, Ying Chen, Wayne~Xin Zhao, Daxiang Dong, Dianhai Yu, Xiangyang
  Zhou, and Lu~Li. 2018.
\newblock Multi-turn response selection for chatbots with deep attention
  matching network.
\newblock In \emph{ACL}.

\bibitem[{Wu et~al.(2016)Wu, Wu, Xing, Zhou, and Li}]{wu2016sequential}
Yu~Wu, Wei Wu, Chen Xing, Ming Zhou, and Zhoujun Li. 2016.
\newblock Sequential matching network: A new architecture for multi-turn
  response selection in retrieval-based chatbots.
\newblock \emph{arXiv preprint arXiv:1612.01627}.

\bibitem[{Xiong et~al.(2017)Xiong, Dai, Callan, Liu, and Power}]{xiong2017end}
Chenyan Xiong, Zhuyun Dai, Jamie Callan, Zhiyuan Liu, and Russell Power. 2017.
\newblock End-to-end neural ad-hoc ranking with kernel pooling.
\newblock In \emph{SIGIR}, pages 55--64. ACM.

\bibitem[{Yang et~al.(2018)Yang, Hu, Dyer, Xing, and
  Berg-Kirkpatrick}]{yang2018unsupervised}
Zichao Yang, Zhiting Hu, Chris Dyer, Eric~P Xing, and Taylor Berg-Kirkpatrick.
  2018.
\newblock Unsupervised text style transfer using language models as
  discriminators.
\newblock In \emph{NeurIPS}.

\bibitem[{Yarats and Lewis(2018)}]{yarats2017hierarchical}
Denis Yarats and Mike Lewis. 2018.
\newblock Hierarchical text generation and planning for strategic dialogue.
\newblock In \emph{ICML}.

\bibitem[{Young et~al.(2007)Young, Schatzmann, Weilhammer, and
  Ye}]{young2007hidden}
Stephanie Young, Jost Schatzmann, Karl Weilhammer, and Hui Ye. 2007.
\newblock The hidden information state approach to dialog management.
\newblock In \emph{Acoustics, Speech and Signal Processing, 2007. ICASSP 2007.
  IEEE International Conference on}, volume~4, pages IV--149. IEEE.

\bibitem[{Zhang et~al.(2018)Zhang, Dinan, Urbanek, Szlam, Kiela, and
  Weston}]{zhang2018personalizing}
Saizheng Zhang, Emily Dinan, Jack Urbanek, Arthur Szlam, Douwe Kiela, and Jason
  Weston. 2018.
\newblock Personalizing dialogue agents: I have a dog, do you have pets too?
\newblock \emph{arXiv preprint arXiv:1801.07243}.

\bibitem[{Zhao et~al.(2019)Zhao, Xie, and Eskenazi}]{zhao2019rethinking}
Tiancheng Zhao, Kaige Xie, and Maxine Eskenazi. 2019.
\newblock Rethinking action spaces for reinforcement learning in end-to-end
  dialog agents with latent variable models.
\newblock \emph{arXiv preprint arXiv:1902.08858}.

\bibitem[{Zhao et~al.(2017)Zhao, Zhao, and Eskenazi}]{zhao2017learning}
Tiancheng Zhao, Ran Zhao, and Maxine Eskenazi. 2017.
\newblock Learning discourse-level diversity for neural dialog models using
  conditional variational autoencoders.
\newblock In \emph{ACL}.

\bibitem[{Zhou et~al.(2018)Zhou, Gao, Li, and Shum}]{zhou2018design}
Li~Zhou, Jianfeng Gao, Di~Li, and Heung-Yeung Shum. 2018.
\newblock The design and implementation of xiaoice, an empathetic social
  chatbot.
\newblock \emph{arXiv preprint arXiv:1812.08989}.

\bibitem[{Zhou et~al.(2016)Zhou, Dong, Wu, Zhao, Yu, Tian, Liu, and
  Yan}]{Zhou2016MultiviewRS}
Xiangyang Zhou, Daxiang Dong, Hua Wu, Shiqi Zhao, Dianhai Yu, Hao Tian, Xuan
  Liu, and Rui Yan. 2016.
\newblock Multi-view response selection for human-computer conversation.
\newblock In \emph{EMNLP}.

\end{thebibliography}
\bibliographystyle{acl_natbib}
\end{document}


\title{Supplementary Material for Target-Guided Open-domain Conversation}


\maketitle
\section{Keyword Extraction}

Following the previous work ~\cite{wang2018chat,yao2018chat} on conversation keyword extraction, we develop a TF-IDF based keyword extractor. For a conversation $\{u_1,u_2,...,u_n\}$, our topic extractor will obtain a keyword list $\{t_1,t_2,...t_k\}$ from each utterance $u_i$ by the following steps: (1) We regard each utterance in a conversation as a document and each word as a term to calculate the TF-IDF value of each word. (2) We ignore the words appearing less than 10 times in all corpus, or have been mentioned in the former utterance of the current conversation. (3) Considering that the importance of each part of speech in a sentence is different, we set different weights to distinguish them. The part-of-speech weights of nouns, verbs, adjectives and others are 2, 1, 0.5 and 0 respectively. (4) We multiply the TF-IDF value with the part-of-speech weight to obtain the composite score of each word, and delete the words whose score is below the given threshold. We consider the words left as the keywords of an utterance in an conversation.

~\ref

\clearpage
\bibliography{acl2019}
\bibliographystyle{acl_natbib}